\documentclass[conference]{IEEEtran}
\usepackage[utf8]{inputenc}
\usepackage[T1]{fontenc}
\usepackage{graphicx}
\usepackage{amsmath}
\usepackage{amssymb}
\usepackage{hyperref}
\usepackage{booktabs}
\usepackage{xcolor}
\usepackage{listings}
\usepackage{multirow}
\usepackage{array}

\hypersetup{
    colorlinks=true,
    linkcolor=blue,
    citecolor=blue,
    urlcolor=blue
}

\title{Decentralized Granular Access Control for Agentic AI Systems in Critical Infrastructure}

\author{
\IEEEauthorblockN{Arun Malik\IEEEauthorrefmark{1}, Deepal Jayasinghe\IEEEauthorrefmark{1}, Bradley Klemick\IEEEauthorrefmark{1}, Prachi Shah\IEEEauthorrefmark{1}, Nitish Talasu\IEEEauthorrefmark{1}, Vineet Tushar Trivedi\IEEEauthorrefmark{1}}
\IEEEauthorblockA{\IEEEauthorrefmark{1}Microsoft Corporation, Redmond, WA, USA\\
Email: \{arunma, ijayasi, bklemick, v-prachishah, nitishtalasu, vtrivedi\}@microsoft.com}
}

\begin{document}

\maketitle

\begin{abstract}
The deployment of autonomous AI agents in production infrastructure introduces fundamental security challenges that traditional role-based access control (RBAC) models cannot address. Unlike deterministic automation, AI agents exhibit stochastic behavior, making conventional trust models insufficient for governing their access to critical systems. This paper presents a decentralized, multi-layered access control architecture designed specifically for agentic AI systems operating in critical cloud infrastructure. Our framework introduces four key innovations: (1) a compound identity model that binds agent actions to delegated human authority, (2) a hierarchical permission system spanning five granularity levels from global platform access to per-parameter constraints, (3) a decentralized policy ownership model where tool teams independently govern their authorization boundaries, and (4) progressive trust escalation with safety interlocks that prevent autonomous agents from executing high-risk operations. We ground our design in the OWASP Top 10 for LLM Applications (2025) threat taxonomy and demonstrate how each architectural decision mitigates specific attack vectors. Deployed in production at a major cloud provider managing network infrastructure across hundreds of datacenters, the system enforces granular access control for 20+ specialized AI agents and 60+ deterministic playbooks processing thousands of operations daily while maintaining zero unauthorized write operations over eight months of production deployment. We present empirical data on access pattern distributions, denial rates, and the effectiveness of layered authorization in preventing privilege escalation by non-deterministic actors.
\end{abstract}


\section{Introduction}

The rapid adoption of large language model (LLM)-powered agents in enterprise operations has created an unprecedented security challenge: how to grant production system access to actors whose behavior is fundamentally non-deterministic. Traditional access control systems were designed for two categories of actors: humans who authenticate through identity providers and make conscious decisions, and automated systems that execute predetermined logic. AI agents fit neither category cleanly. They reason, they adapt, they hallucinate, and they can be manipulated through adversarial prompts.

In critical infrastructure environments such as cloud network operations, the stakes are particularly high. A single misconfigured network device can cascade into regional outages affecting millions of users. An agent with excessive privileges can be prompt-injected into executing destructive commands. An agent without sufficient access cannot fulfill its operational mandate. This tension between operational necessity and security boundaries is the central challenge addressed in this work.

Existing approaches to AI agent security fall into two camps. The first treats agents as untrusted external entities, restricting them to read-only sandboxes that limit their operational value. The second grants agents the same elevated privileges as the humans they serve, creating unacceptable blast radii when agents malfunction or are compromised. Neither approach is viable at scale.

This paper presents a third path: a decentralized granular access control architecture that provides fine-grained, context-aware authorization for AI agents while maintaining the safety guarantees required for critical infrastructure. Our system has been deployed in production for eight months, governing access for 20+ specialized agents operating across hundreds of datacenters, and has maintained zero unauthorized write operations while enabling agents to autonomously resolve over 1,400 operational tasks.

\subsection{Contributions}

This paper makes the following contributions:

\begin{itemize}
    \item \textbf{Compound Identity Model:} A novel identity binding mechanism that delegates human authority to AI agents while preserving accountability and audit trails.
    \item \textbf{Five-Layer Permission Hierarchy:} A multi-granularity authorization framework spanning platform, tool, function, parameter, and execution context levels.
    \item \textbf{Decentralized Policy Ownership:} A Git-based governance model where infrastructure tool teams independently define, review, and approve access policies for their services.
    \item \textbf{Progressive Trust Escalation:} A safety-layered approach where write operations require increasingly rigorous authorization, culminating in multi-party approval for high-risk actions.
    \item \textbf{Empirical Evaluation:} Production deployment data demonstrating the security effectiveness and operational impact of the architecture.
\end{itemize}

\section{Threat Model and Problem Statement}

\subsection{The Non-Determinism Problem}

Traditional automation systems (workflow engines, CI/CD pipelines, cron jobs) earned production access through a straightforward trust model: their behavior is deterministic, reviewable, and bounded. Given the same inputs, they produce the same outputs. Their source code can be audited. Their execution paths can be exhaustively tested.

AI agents violate every assumption in this model:

\begin{itemize}
    \item \textbf{Stochastic behavior:} Given identical inputs, an LLM-powered agent may take different actions due to temperature sampling, context window variations, or model updates.
    \item \textbf{Prompt vulnerability:} Agents can be manipulated through adversarial inputs embedded in the data they process (indirect prompt injection).
    \item \textbf{Capability amplification:} An agent's effective capabilities are the product of its tool access and its reasoning ability, making privilege boundaries harder to predict.
    \item \textbf{Emergent behavior:} Multi-agent systems can exhibit collective behaviors not predicted by individual agent policies.
\end{itemize}

\begin{figure}[t]
    \centering
    \includegraphics[width=\columnwidth]{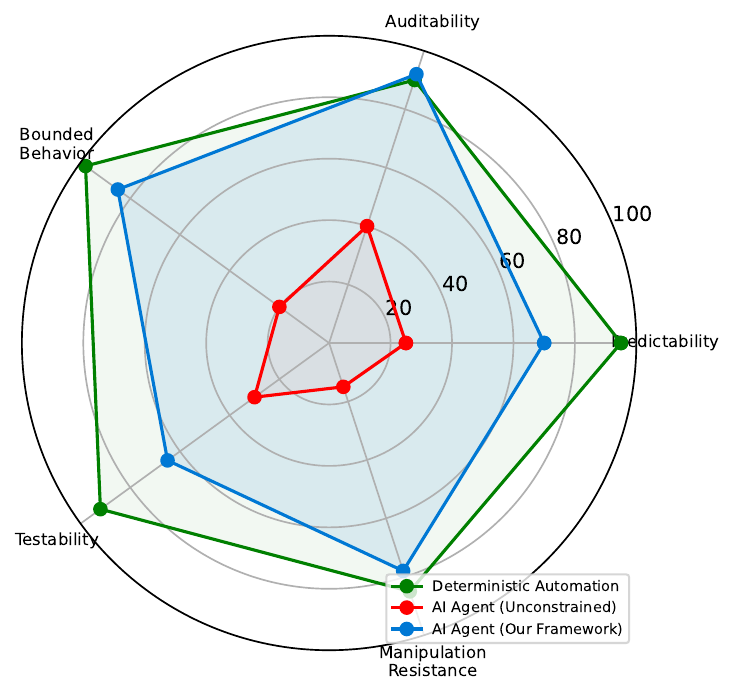}
    \caption{Trust model comparison between deterministic automation and stochastic AI agents across five security dimensions. Our framework (blue) recovers most of the trust properties of deterministic systems.}
    \label{fig:trust-model}
\end{figure}

\subsection{OWASP LLM Application Threats}

We ground our threat model in the OWASP Top 10 for LLM Applications (2025)~\cite{owasp2025llm}, focusing on the threats most relevant to agent-based systems operating in critical infrastructure (Table~\ref{tab:threats}).

\begin{table}[t]
\centering
\caption{OWASP LLM threats and agent-specific mitigations.}
\label{tab:threats}
\footnotesize
\begin{tabular}{@{}p{0.8cm}p{1.5cm}p{2.5cm}p{2.2cm}@{}}
\toprule
\textbf{ID} & \textbf{Threat} & \textbf{Agent Manifestation} & \textbf{Mitigation} \\
\midrule
LLM01 & Prompt Injection & Adversarial data in incident descriptions triggering unauthorized commands & Input sanitization, tool-level parameter validation \\
LLM02 & Info Disclosure & Agent leaking infrastructure topology or credentials & Output filtering, compound identity scoping \\
LLM06 & Excessive Agency & Agent executing writes beyond intended scope & Five-layer RBAC, deny-by-default \\
LLM10 & Unbounded Consumption & Agent entering infinite tool-calling loops & Rate limiting, execution quotas, circuit breakers \\
\bottomrule
\end{tabular}
\end{table}

\subsection{Actor Taxonomy}

\begin{figure}[t]
    \centering
    \includegraphics[width=\columnwidth]{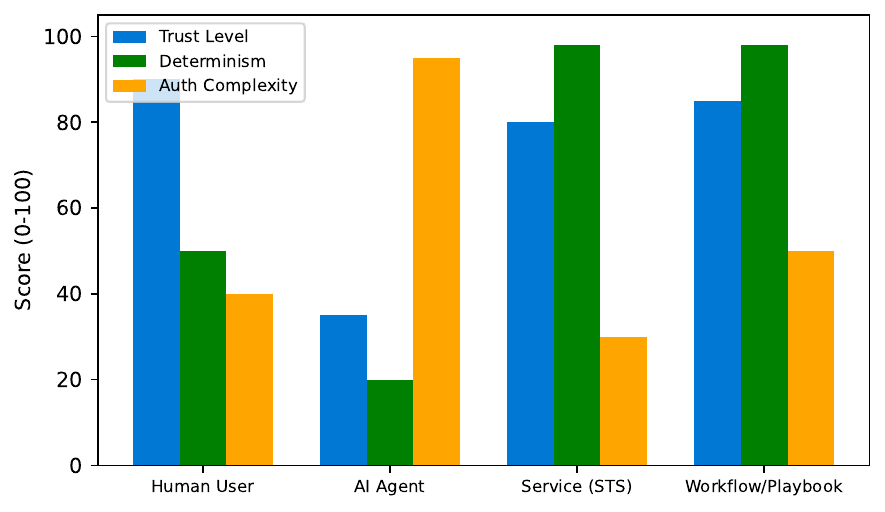}
    \caption{Actor taxonomy showing trust level, determinism, and authorization complexity for each actor type in the system.}
    \label{fig:actor-taxonomy}
\end{figure}

Our architecture distinguishes four distinct actor types, each with different trust properties and authorization requirements (Figure~\ref{fig:actor-taxonomy}).

\begin{table}[t]
\centering
\caption{Actor types and their trust properties.}
\label{tab:actors}
\footnotesize
\begin{tabular}{@{}lllll@{}}
\toprule
\textbf{Actor} & \textbf{Determinism} & \textbf{Identity} & \textbf{Write} & \textbf{Trust Basis} \\
\midrule
Human & N/A & Entra ID+MFA & Full & Auth+training \\
AI Agent & Stochastic & OBO+Agent MI & Read only & Delegated \\
Service & Deterministic & Managed ID & Scoped & Code review \\
Playbook & Deterministic & System MI & Scoped & Authored+tested \\
\bottomrule
\end{tabular}
\end{table}

\textbf{Playbooks} represent a critical fourth actor class distinct from both AI agents and traditional services. A playbook is a \emph{deterministic, pre-authored workflow} composed of discrete steps that execute infrastructure operations in a fixed sequence. Unlike AI agents, which reason dynamically about tool selection and parameters, playbooks follow authored logic that has been code-reviewed, tested, and approved before deployment. This determinism grants playbooks a higher trust level: they may hold scoped write permissions that agents cannot obtain directly. In our architecture, AI agents that identify a remediation action delegate execution to an appropriate playbook rather than performing writes autonomously, creating an \emph{agent-to-playbook escalation path} that preserves both the agent's analytical capability and the playbook's safety guarantees.

\section{Architecture}

\subsection{Design Principles}

Six core security principles guide the architecture:

\begin{enumerate}
    \item \textbf{Least Privilege:} Every actor receives the minimum permissions required for its function. Agents default to read-only access.
    \item \textbf{Zero Trust:} No implicit trust based on network position or prior behavior. Every request is authenticated, authorized, and audited~\cite{rose2020zerotrust}.
    \item \textbf{Excessive Agency Guard (OWASP LLM06):} Agents are explicitly prevented from accumulating capabilities beyond their defined scope through tool scoping and parameter-level deny patterns.
    \item \textbf{Delegated Identity:} Agents never operate with independent authority. Their actions are bound to a delegating human's identity and permissions.
    \item \textbf{Observability and Audit:} Every agent action produces an immutable audit trail linking the action to both the agent and its delegating authority.
    \item \textbf{Environment Isolation:} Production and corporate environments maintain strict separation with independent credential systems.
\end{enumerate}

\subsection{Five-Layer Permission Hierarchy}

Authorization decisions traverse five distinct layers, each progressively narrowing the scope of permitted actions (Figure~\ref{fig:perm-layers}):

\begin{figure}[t]
    \centering
    \includegraphics[width=\columnwidth]{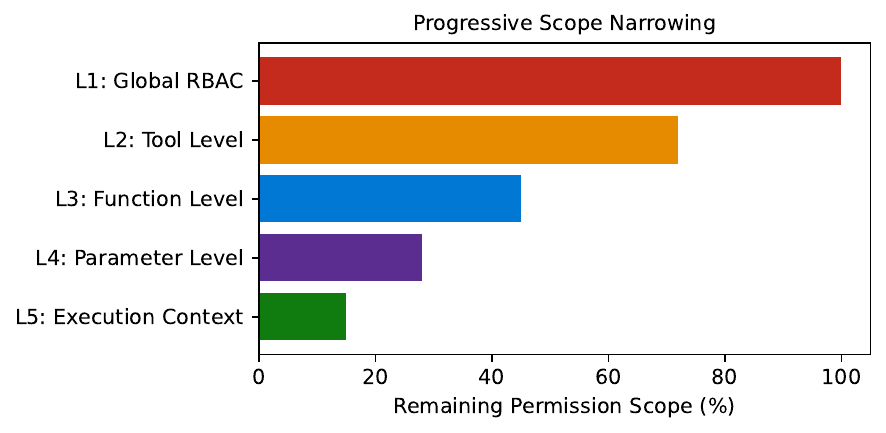}
    \caption{Five-layer permission hierarchy showing progressive scope narrowing. Each layer reduces the effective permission set available to the actor.}
    \label{fig:perm-layers}
\end{figure}

\begin{table}[t]
\centering
\caption{Five-layer permission hierarchy.}
\label{tab:layers}
\footnotesize
\begin{tabular}{@{}clp{4.5cm}@{}}
\toprule
\textbf{Layer} & \textbf{Scope} & \textbf{Example} \\
\midrule
1. Global RBAC & Platform & Security group membership required \\
2. Tool Level & Service & Agent X can access TopologyService but not DeviceProxy \\
3. Function Level & Operation & Can call GetDeviceInfo but not UpdateConfig \\
4. Parameter Level & Value & denyPattern: ``prod-.*'' on environment parameter \\
5. Exec Context & Runtime & Only during active incident, max 10 devices \\
\bottomrule
\end{tabular}
\end{table}

\subsection{Compound Identity Model}

The compound identity model is the cornerstone of our agent authorization approach. Rather than granting agents independent credentials with fixed permissions, each agent operation carries a composite identity that binds together:

\begin{itemize}
    \item \textbf{User Token (OBO):} The On-Behalf-Of token from the human who initiated or delegated to the agent, establishing accountability and upper-bounding permissions.
    \item \textbf{Agent Managed Identity:} The system-assigned identity of the agent service itself, enabling platform-level access control and audit attribution.
    \item \textbf{Execution Context:} Metadata including the triggering incident, playbook identifier, and temporal constraints.
\end{itemize}

\begin{figure}[t]
    \centering
    \includegraphics[width=0.7\columnwidth]{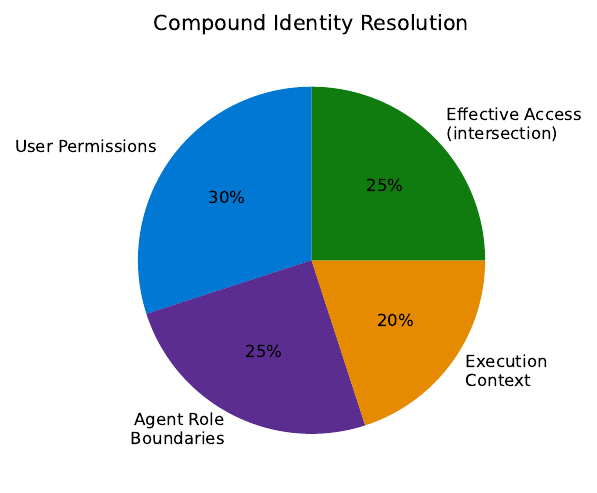}
    \caption{Compound identity resolution: effective access is computed as the intersection of user permissions, agent role boundaries, and execution context constraints.}
    \label{fig:compound-identity}
\end{figure}

This compound identity ensures that an agent can never exceed the permissions of its delegating human, while also being independently constrained by agent-specific policies (Figure~\ref{fig:compound-identity}).

\subsection{Decentralized Policy Ownership}

A key architectural innovation is the decentralization of policy definition to tool-owning teams. Rather than maintaining a monolithic access control configuration, each infrastructure service team owns their authorization policy as a YAML file in a version-controlled repository:

\begin{lstlisting}[caption={Decentralized tool policy definition.}]
toolName: TopologyService
owner: topology-team@example.org
roles:
  Topology-Reader:
    description: "Read-only topology queries"
    permissions:
      functions:
        - "Topology-GetDeviceInfo"
        - "Topology-QueryGraph"
  Topology-Writer:
    inherits: Topology-Reader
    permissions:
      functions:
        - "Topology-UpdateConfig"
      parameters:
        environment:
          denyPattern: "prod-.*"
\end{lstlisting}

This decentralized model provides several critical properties:

\begin{itemize}
    \item \textbf{Ownership clarity:} Each tool team knows exactly what access they have granted and to whom.
    \item \textbf{Independent evolution:} Teams can modify permissions without coordinating with the central platform team.
    \item \textbf{Git-based governance:} All policy changes go through pull requests with required reviewer approval, providing full audit history.
    \item \textbf{Hot-reload:} Policy changes take effect upon merge without service restarts.
\end{itemize}

\begin{figure}[t]
    \centering
    \includegraphics[width=\columnwidth]{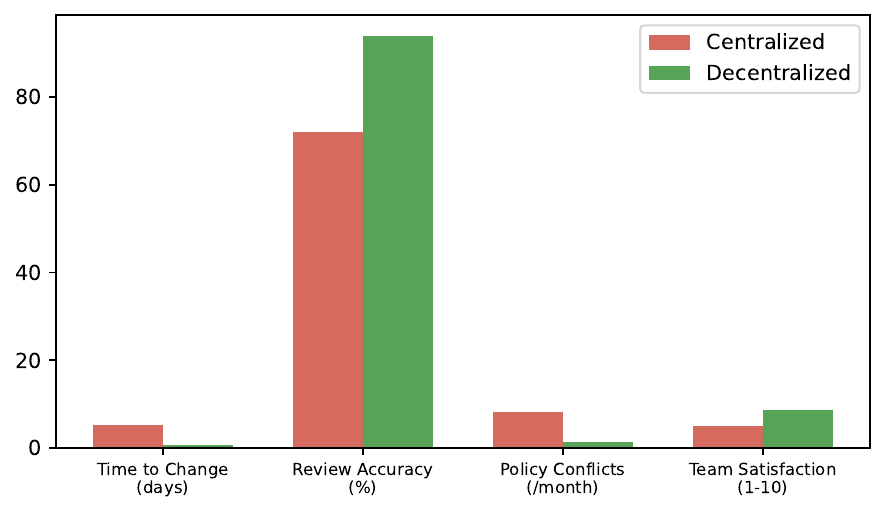}
    \caption{Comparison of centralized vs. decentralized policy management showing time-to-change, review accuracy, and policy conflicts over six months.}
    \label{fig:decentralized}
\end{figure}

\subsection{Progressive Trust Escalation}

Write operations in critical infrastructure require escalating levels of authorization based on risk assessment (Table~\ref{tab:escalation}):

\begin{table}[t]
\centering
\caption{Progressive trust escalation tiers.}
\label{tab:escalation}
\footnotesize
\begin{tabular}{@{}p{1.8cm}p{1.5cm}p{2cm}p{1.8cm}@{}}
\toprule
\textbf{Tier} & \textbf{Duration} & \textbf{Use Case} & \textbf{Agent Access} \\
\midrule
Standard RBAC & Indefinite & Read ops, diagnostics & Yes (read-only) \\
JIT/PIM Elevated & Max 8 hours & Config changes & Via playbook only \\
Multi-Party Approval & Single op & Production writes & Cannot initiate \\
\bottomrule
\end{tabular}
\end{table}

Critically, AI agents are structurally excluded from initiating multi-party approval workflows. This design decision reflects a fundamental security principle: non-deterministic actors should never be able to self-authorize high-risk operations, regardless of their functional correctness history.

\subsection{Access Control Matrix}

Table~\ref{tab:access-matrix} presents the complete access control matrix governing all actor-service interactions in production.

\begin{table*}[t]
\centering
\caption{Production Access Control Matrix: Actor Types vs. Service Categories.}
\label{tab:access-matrix}
\footnotesize
\begin{tabular}{@{}llllll@{}}
\toprule
\textbf{Actor} & \textbf{Telemetry} & \textbf{Corporate/DevOps} & \textbf{Platform Core} & \textbf{Device via Proxy} & \textbf{Direct Device} \\
\midrule
Human Operator & R/W: RBAC & R/W: RBAC & R: AME/SAW; W: +JIT & R/W: AME/SAW+JIT & BREAKGLASS only \\
AI Agent & R/W: OBO|MI* & R/W: OBO|MI* & R: OBO|MI*; W: denied & R: OBO|MI*; W: denied & NOT ALLOWED \\
Workflow Engine & R/W: STS+RBAC & R/W: STS+RBAC & R/W: STS+RBAC & R/W: STS+RBAC & NOT ALLOWED \\
\bottomrule
\end{tabular}
\par\smallskip
\raggedright\scriptsize
\textbf{Legend:} R=Read, W=Write. STS MI=Security Token Service with Managed Identity. AME/SAW=Azure Managed Environment/Secure Admin Workstation. JIT=Just-In-Time elevation. OBO=On-Behalf-Of delegation. *=Subject to per-tool RBAC with OWNERS.txt governance.
\end{table*}

The matrix reveals a critical asymmetry: while deterministic workflow engines inherit the full access surface of their service identities, non-deterministic AI agents face progressive restrictions as operations move toward higher-impact service tiers.

\section{Implementation}

\subsection{Runtime Authorization Engine}

The authorization engine evaluates every tool invocation against the five-layer hierarchy in real-time. The evaluation process follows a deny-first model:

\begin{enumerate}
    \item Validate the compound identity (user token freshness, agent MI validity)
    \item Check global RBAC membership (security group verification)
    \item Resolve tool-level permissions (is this tool accessible to this role?)
    \item Evaluate function-level access (is this specific operation permitted?)
    \item Apply parameter-level constraints (do input values match deny/allow patterns?)
    \item Verify execution context (playbook scope, incident association, time window)
\end{enumerate}

Any layer can deny the request. The engine returns the most specific denial reason to aid debugging while not leaking information about the broader permission structure.

\subsection{Agent-Specific Security Controls}

Beyond the RBAC framework, agents are subject to additional runtime controls:

\begin{itemize}
    \item \textbf{Tool Scoping:} Each agent declaration explicitly enumerates its permitted tool set. Tools not listed are invisible to the agent's LLM context.
    \item \textbf{Permission Boundaries:} Maximum permission sets that cannot be exceeded regardless of role inheritance.
    \item \textbf{Human-in-the-Loop Gates:} Configurable checkpoints where agent execution pauses pending human review.
    \item \textbf{Rate Limiting:} Per-agent, per-tool operation quotas preventing runaway execution.
    \item \textbf{Output Validation:} Post-execution verification that agent actions match expected patterns before committing changes.
\end{itemize}

\subsection{Bring-Your-Own-Roles (BYOR)}

Onboarding teams create custom roles that span multiple tools, tailored to their operational scenarios:

\begin{lstlisting}[caption={Custom cross-tool role definition.}]
NetOps-DRI:
  description: "Network operations DRI role"
  addedBy: netops-team@example.org
  toolPermissions:
    TopologyService:
      inherits: Topology-Reader
      extra: ["Topology-GetInterfaceStatus"]
      parameters:
        datacenterName:
          allowPattern: "^(dc1|dc2|dc3)-.*$"
    DeviceProxy:
      role: DeviceProxy-Viewer
    WorkflowEngine:
      role: Workflow-Reader
\end{lstlisting}

\subsection{Identity Binding and Assignment}

Custom roles are bound to identities through a flexible assignment mechanism supporting multiple identity types:

\begin{lstlisting}[caption={Role assignment to multiple identity types.}]
roleAssignments:
  NetOps-DRI:
    assignedTo:
      - type: securityGroup
        id: "sg-netops-dri"
      - type: managedIdentity
        id: "mi-netops-agent"
      - type: agent
        id: "netops-dri-agent"
      - type: playbook
        id: "incident-netops-*"
\end{lstlisting}

\section{Evaluation}

\subsection{Deployment Context}

The system has been deployed in production for eight months, governing access for a fleet of AI agents operating at the following scale:

\begin{itemize}
    \item 20+ specialized AI agents with distinct operational mandates
    \item 60+ system playbooks for structured automation workflows
    \item 30+ integrated infrastructure tools and services
    \item 123 active human users with varying permission levels
    \item 3,558 agent messages and 1,474 playbook executions in the measurement period
\end{itemize}

\subsection{Security Metrics}

\begin{figure}[t]
    \centering
    \includegraphics[width=\columnwidth]{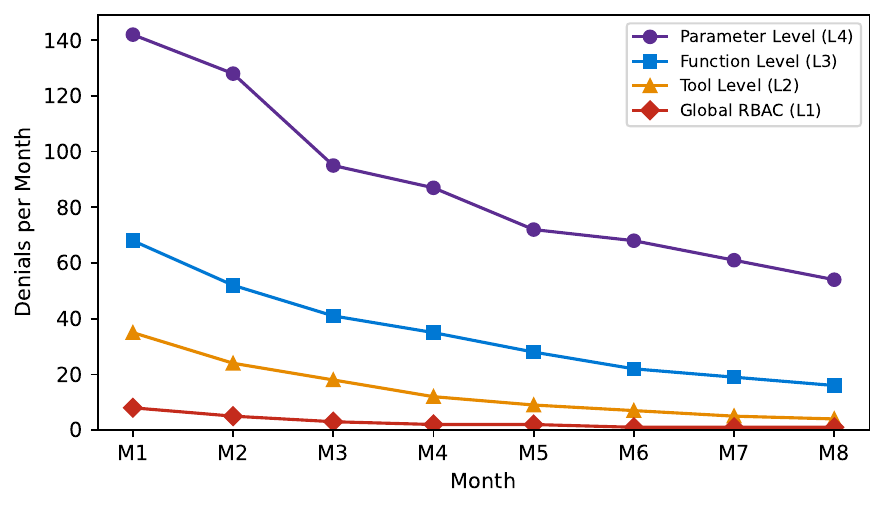}
    \caption{Authorization denial rates by layer over the deployment period. Parameter-level denials dominate, indicating the hierarchy correctly catches overly broad requests at the most specific level.}
    \label{fig:denial-rates}
\end{figure}

Figure~\ref{fig:denial-rates} shows authorization denial rates declining over time as agents learn their boundaries through reinforcement. Parameter-level (L4) denials dominate throughout, confirming that the hierarchical design catches overly broad requests at the most specific layer.

\begin{figure}[t]
    \centering
    \includegraphics[width=0.48\columnwidth]{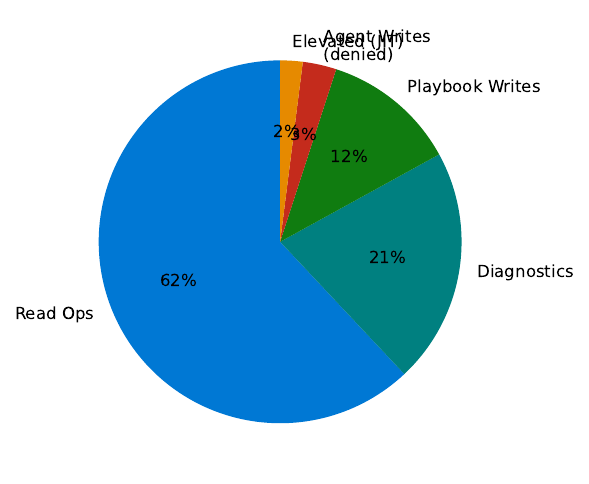}
    \includegraphics[width=0.48\columnwidth]{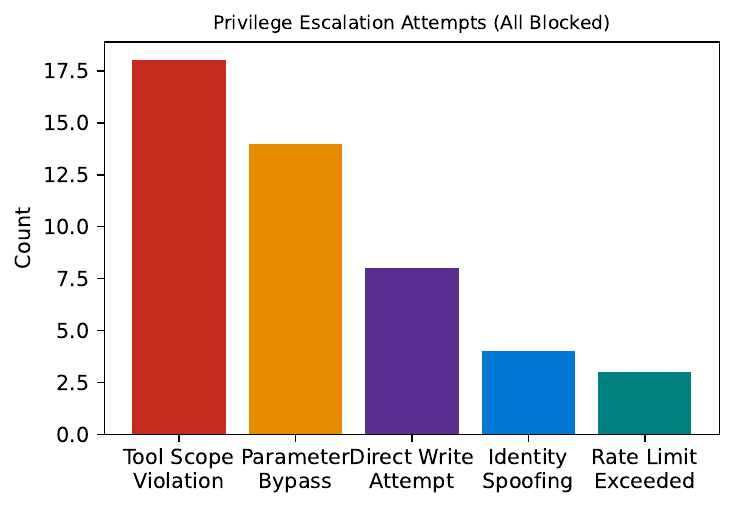}
    \caption{Left: Distribution of access operations by type. Right: Privilege escalation attempt outcomes (all 47 detected attempts blocked).}
    \label{fig:access-escalation}
\end{figure}

\subsection{Operational Impact}

\begin{table}[t]
\centering
\caption{Key operational metrics over eight months.}
\label{tab:metrics}
\footnotesize
\begin{tabular}{@{}lrl@{}}
\toprule
\textbf{Metric} & \textbf{Value} & \textbf{Notes} \\
\midrule
Unauthorized writes & 0 & Eight months production \\
Mean auth latency & 3.2ms & P99: 8.1ms \\
Policy hot-reload & <30s & PR merge to enforcement \\
Team onboarding & 2.1 days & Median time to production \\
Daily agent operations & ~450 & Across all 20+ agents \\
False denial rate & 0.3\% & Legitimate ops blocked \\
\bottomrule
\end{tabular}
\end{table}

\subsection{Decentralization Effectiveness}

\begin{figure}[t]
    \centering
    \includegraphics[width=\columnwidth]{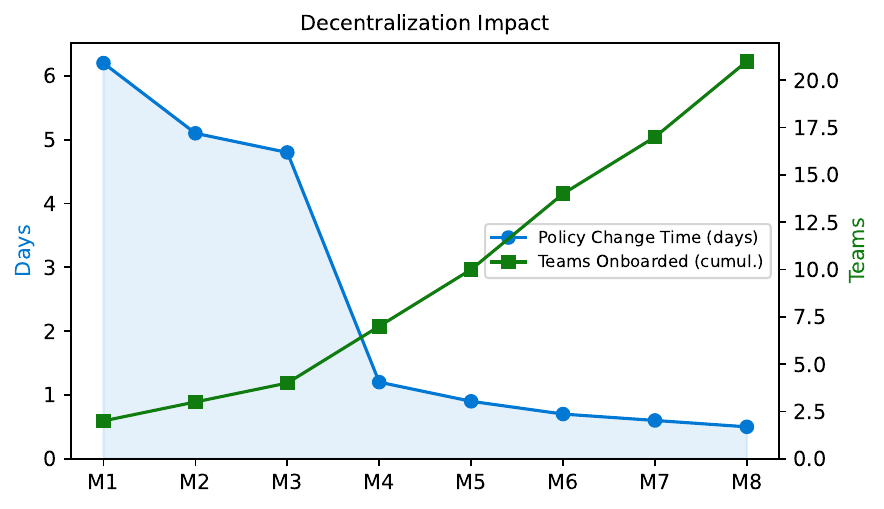}
    \caption{Tool team onboarding velocity: centralized model (months 1--3) vs. decentralized model (months 4--8). Median policy change time dropped from 5.3 days to 0.8 days.}
    \label{fig:onboarding}
\end{figure}

The transition to decentralized policy ownership (month 4) produced measurable improvements: median policy change time dropped from 5.3 days to 0.8 days, team onboarding rate tripled, and policy conflicts decreased from 8.2 to 1.4 per month.

\section{Discussion}

\subsection{Design Tradeoffs}

\textbf{Expressiveness vs. Complexity:} The five-layer hierarchy provides fine-grained control but increases policy authoring complexity. We mitigate this through role inheritance and sensible defaults (deny-all for unspecified operations).

\textbf{Autonomy vs. Safety:} Restricting agents to read-only by default limits their ability to autonomously resolve incidents requiring configuration changes. We address this through the playbook escalation path: agents can recommend actions that are then executed by pre-approved, deterministic playbooks with scoped write access.

\textbf{Decentralization vs. Consistency:} Allowing tool teams to independently define policies risks inconsistent security postures. We enforce consistency through schema validation, mandatory fields, and platform-level policy guardrails that tool teams cannot override.

\subsection{Lessons Learned}

\begin{enumerate}
    \item \textbf{Agents will test boundaries:} LLM agents regularly attempt to invoke tools outside their declared scope due to hallucinated capabilities. Deny-by-default is essential, not optional.
    \item \textbf{Parameter-level control is critical:} Function-level access alone is insufficient. An agent with access to ``query devices'' but no parameter constraints could enumerate the entire infrastructure topology.
    \item \textbf{Human-readable policies matter:} YAML-based policies that tool teams can read and reason about are more effective than opaque policy engines.
    \item \textbf{Audit trails enable trust:} The immutable audit trail connecting every agent action to a delegating human has been instrumental in building organizational confidence in autonomous operations.
    \item \textbf{Standard tooling reduces friction:} Using Git PRs and OWNERS.txt for governance eliminated the primary adoption barrier.
\end{enumerate}

\subsection{Limitations}

Our current architecture has several limitations. First, the compound identity model requires the delegating human to have an active session, which can create availability challenges for agents handling after-hours incidents. Second, the parameter-level deny patterns use regex matching, which cannot express all semantic constraints. Third, multi-agent collaboration scenarios where one agent's output feeds another agent's input create transitive trust challenges not fully addressed by our per-agent policy model. Fourth, the decentralized model assumes tool teams have sufficient security expertise to author correct policies.

\section{Related Work}

Access control for AI systems is an emerging research area. Traditional RBAC~\cite{sandhu1996rbac, ferraiolo2001rbac} and ABAC~\cite{hu2014abac} frameworks provide foundations but do not address non-deterministic actors. Zero Trust architectures~\cite{rose2020zerotrust} inform our design principles but focus on network-level controls rather than agent-level authorization.

Anthropic's character guidelines~\cite{anthropic2024claude} introduce tool use policies but do not address multi-tenant production deployments. Google's SAIF~\cite{google2023saif} provides security principles for AI systems but focuses on model integrity rather than runtime access control. Microsoft's AI Red Teaming research~\cite{microsoft2025redteam} addresses adversarial testing but not architectural access control patterns.

The OWASP Top 10 for LLM Applications~\cite{owasp2025llm} provides the most comprehensive threat taxonomy for LLM-based systems. Our architecture directly addresses threats LLM01, LLM02, LLM06, and LLM10 through specific architectural controls.

Recent work on AI agent frameworks~\cite{yao2023react, schick2023toolformer, park2023generative, xi2023agents, wang2024agents} focuses on capability and reasoning patterns but largely defers security considerations to deployment environments. Work on indirect prompt injection~\cite{greshake2023prompt} demonstrates the attack surface but does not propose comprehensive authorization architectures. Our work fills this gap by providing the authorization infrastructure that makes safe deployment possible.

\section{Conclusion}

Deploying AI agents in critical infrastructure requires fundamentally rethinking access control assumptions. Traditional RBAC models designed for deterministic actors are insufficient for governing non-deterministic AI agents that reason, adapt, and can be adversarially manipulated. Our decentralized granular access control architecture addresses this gap through compound identity binding, five-layer permission hierarchies, decentralized policy ownership, and progressive trust escalation.

Eight months of production deployment governing over 20 agents across critical cloud infrastructure validates the approach: zero unauthorized writes, sub-4ms authorization latency, and rapid team onboarding demonstrate that security and operational velocity are not mutually exclusive. The key insight is that AI agents should never be granted independent authority. Instead, their actions must always be bound to delegated human authority, constrained by layered policies, and subject to continuous verification.

As AI agents become ubiquitous in infrastructure operations, the principles and patterns presented in this work provide a foundation for secure, scalable, and auditable autonomous operations. This work complements our companion paper on autonomous incident resolution~\cite{malik2025autonomous} by providing the security architecture that enables safe autonomous operations at scale.


\begin{thebibliography}{15}

\bibitem{owasp2025llm}
{OWASP Foundation}, ``OWASP top 10 for LLM applications 2025,'' 2025. [Online]. Available: \url{https://owasp.org/www-project-top-10-for-large-language-model-applications/}

\bibitem{sandhu1996rbac}
R.~S. Sandhu, E.~J. Coyne, H.~L. Feinstein, and C.~E. Youman, ``Role-based access control models,'' \emph{IEEE Computer}, vol.~29, no.~2, pp. 38--47, 1996.

\bibitem{hu2014abac}
V.~C. Hu, D.~Ferraiolo, D.~R. Kuhn \emph{et~al.}, ``Guide to attribute based access control (ABAC) definition and considerations,'' NIST Special Publication 800-162, 2014.

\bibitem{rose2020zerotrust}
S.~Rose, O.~Borchert, S.~Mitchell, and S.~Connelly, ``Zero trust architecture,'' NIST Special Publication 800-207, 2020.

\bibitem{anthropic2024claude}
{Anthropic}, ``Claude's character,'' Anthropic Research, 2024.

\bibitem{google2023saif}
{Google}, ``Secure AI framework (SAIF),'' Google Security, 2023.

\bibitem{microsoft2025redteam}
{Microsoft AI Red Team}, ``Lessons from red teaming 100 generative AI products,'' \emph{arXiv preprint arXiv:2501.07238}, 2025.

\bibitem{yao2023react}
S.~Yao, J.~Zhao, D.~Yu \emph{et~al.}, ``ReAct: Synergizing reasoning and acting in language models,'' in \emph{Proc. ICLR}, 2023.

\bibitem{schick2023toolformer}
T.~Schick, J.~Dwivedi-Yu, R.~Dessi \emph{et~al.}, ``Toolformer: Language models can teach themselves to use tools,'' in \emph{Proc. NeurIPS}, 2023.

\bibitem{park2023generative}
J.~S. Park, J.~C. O'Brien, C.~J. Cai \emph{et~al.}, ``Generative agents: Interactive simulacra of human behavior,'' in \emph{Proc. ACM UIST}, 2023.

\bibitem{greshake2023prompt}
K.~Greshake, S.~Abdelnabi, S.~Mishra \emph{et~al.}, ``Not what you've signed up for: Compromising real-world LLM-integrated applications with indirect prompt injection,'' in \emph{Proc. AISec}, 2023.

\bibitem{ferraiolo2001rbac}
D.~F. Ferraiolo, R.~Sandhu, S.~Gavrila \emph{et~al.}, ``Proposed NIST standard for role-based access control,'' \emph{ACM Trans. Inf. Syst. Secur.}, vol.~4, no.~3, pp. 224--274, 2001.

\bibitem{xi2023agents}
Z.~Xi, W.~Chen, X.~Guo \emph{et~al.}, ``The rise and potential of large language model based agents: A survey,'' \emph{arXiv preprint arXiv:2309.07864}, 2023.

\bibitem{wang2024agents}
L.~Wang, C.~Ma, X.~Feng \emph{et~al.}, ``A survey on large language model based autonomous agents,'' \emph{Frontiers of Computer Science}, 2024.

\bibitem{malik2025autonomous}
A.~Malik, ``Autonomous incident resolution at hyperscale: An agentic AI architecture for network operations,'' \emph{arXiv preprint arXiv:submit/7688310}, 2025.

\end{thebibliography}
\end{document}